\documentclass[preprint]{elsarticle}

\usepackage{graphicx}
\usepackage{comment}
\usepackage{amsmath,amssymb} % define this before the line numbering.
\usepackage{color}
\usepackage{algorithm}
\usepackage{algpseudocode}
\usepackage{subcaption}
\usepackage{booktabs} % Allows the use of \toprule, \midrule and \bottomrule in tables for horizontal lines
\usepackage{array}
\usepackage{multirow}

\usepackage[inline]{enumitem} 
\usepackage{xspace}
\newcommand*{\eg}{\emph{e.g.}\@\xspace}
\newcommand*{\ie}{\emph{i.e.}\@\xspace}

\newcolumntype{P}[1]{>{\centering\arraybackslash}p{#1}}

\usepackage{hyperref}
\journal{Pattern Recognition}

\begin{document}

\begin{frontmatter}

\title{Temporally Smooth Online Action Detection using Cycle-consistent Future Anticipation}

%% or include affiliations in footnotes:
\author[yonseiaddress]{Young Hwi Kim}
\ead{younghwikim@yonsei.ac.kr}

\author[yonseiaddress]{Seonghyeon Nam}
\ead{shnnam@yonsei.ac.kr}

\author[yonseiaddress]{Seon Joo Kim\corref{mycorrespondingauthor}}
\cortext[mycorrespondingauthor]{Corresponding author}
\ead{seonjookim@yonsei.ac.kr}

\address[yonseiaddress]{Dept. of Computer Science, Yonsei University, 50 Yonsei-ro, Seodaemun-gu, Seoul, Republic of Korea}

\begin{abstract}
Many video understanding tasks work in the offline setting by assuming that the input video is given from the start to the end.
However, many real-world problems require the online setting, making a decision immediately using only the current and the past frames of videos such as in autonomous driving and surveillance systems.
In this paper, we present a novel solution for online action detection by using a simple yet effective RNN-based networks called the Future Anticipation and Temporally Smoothing network (FATSnet).
The proposed network consists of a module for anticipating the future that can be trained in an unsupervised manner with the cycle-consistency loss, and another component for aggregating the past and the future for temporally smooth frame-by-frame predictions. 
We also propose a solution to relieve the performance loss when running RNN-based models on very long sequences. 
Evaluations on TVSeries, THUMOS'14, and BBDB show that our method achieve the state-of-the-art performances compared to the previous works on online action detection.
\end{abstract}


\end{frontmatter}

%\linenumbers

\section{Introduction}
Success of deep neural networks (DNN) has made a profound impact on computer vision, totally changing the paradigm of how we process and understand images.
This trend has also been adopted in video understanding and most video understanding tasks now employ deep learning, achieving state-of-the-art performances.

Among many tasks in video understanding, action recognition has received the most attention until now. 
The goal of action recognition is to classify a trimmed video sequence into one of the predefined action classes. 
Many deep architectures have been introduced for action recognition \cite{tran2015learning,li2020spatio,hao2019spatiotemporal}, providing the essential building blocks for learning the representation of videos.

\begin{figure}
\begin{center}
\includegraphics[width=0.9\linewidth]{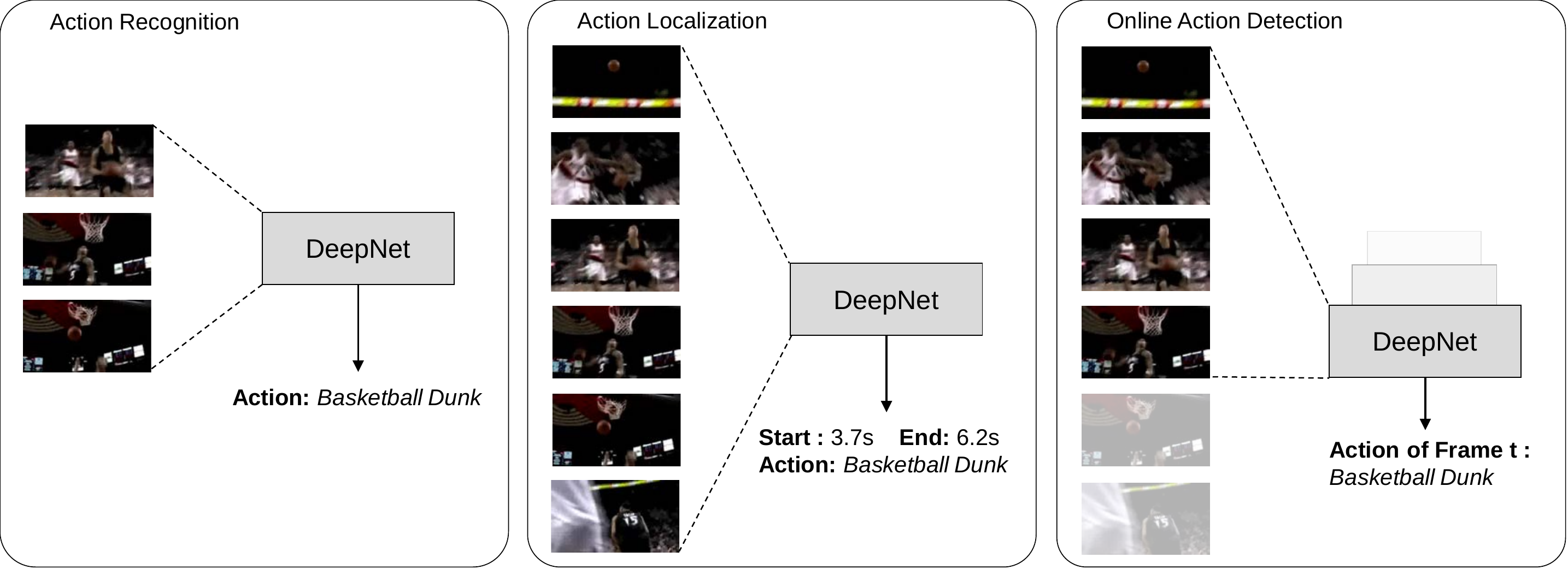}
\end{center}
  \caption{Different tasks in video understanding. We target the online action detection problem in this paper.}
\label{fig:intro}
\end{figure}

Another video understanding task that has received considerable attention is action localization (or offline action detection).
In action localization, the task is to classify the action class with temporal boundaries given untrimmed video sequences. 
Many deep learning algorithms for action localization have been introduced in different settings including supervised~\cite{chao2018rethinking,long2019gaussian,gao2020play} and weakly-supervised~\cite{paul2018wtalc,shou2018autoloc}.

While action recognition and localization are very important tasks for video understanding, it is difficult to apply the solutions to those problems to build online computer vision systems that solve real-time video related problems in practice.
Both problems fall into the offline setting, meaning that the whole video sequence needs to be given from the start to the end as the input.
With offline approaches, actions can be only detected after the fact.
However, many real-world problems have to deal with online streaming video settings where the decision has to be made as an action happens, using only the current and the past frames, not looking at the whole video.
Some examples include computer vision systems for autonomous driving, surveillance systems, and recognition of streaming videos such as sports and movies. 

To this end, online action detection task was introduced in \cite{de2016online} with the goal of
detecting actions and events online.
In the online setting, the actions need to be detected and recognized based on only the current and the past frames (Fig.~\ref{fig:intro}).
One could argue that the online detection problem is more challenging and resembles the human vision system better, compared to the offline settings.
People can recognize actions without observing the whole sequence from the start to the end, based on the minimum observation and from the past experiences. 

Another similar work to online action detection problem is early action detection \cite{hoai2012max,hoai2014max}.
In early action detection, the goal is to recognize an action as early as possible after looking at only a part of the video sequence.
For this specific problem, the action class is given as a prior and the system only has to detect the the action as early as possible, after it starts but before it ends.
While the works in \cite{hoai2012max,hoai2014max} were under  simplified settings (\eg, one action instance per video), \cite{gao2019startnet,shou2018online} defined a new task named online detection of action start (ODAS).
They extended the early action detection problem to detecting on realistic videos that are unconstrained and contain complex backgrounds of large variety.

In this paper, we propose a simple yet effective RNN-based networks, Future Anticipation and Temporally Smoothing network (FATSnet), for the online action detection task.
A module for anticipating the future that can be trained in an unsupervised manner with a novel loss function is combined with an aggregation module for the online detection of actions.
We also define a new temporal smoothing method that fits better with the online settings, as well as a technique to run RNN-based models on very long sequences.
Evaluations on TVSeries~\cite{de2016online}, THUMOS'14~\cite{THUMOS14}, and BBDB~\cite{shim2018teaching} show that our method achieve the state-of-the-art performances compared to the previous works on online detection.

In Section \ref{Sec:Related}, we begin by reviewing previous works on online detection and summarizing the novel contributions of our work compared to the previous methods.
We explain our algorithm in details in Section ~\ref{Sec:Approach}, followed by experiments in Section ~\ref{sec:experiments}.
We conclude the paper with a discussion in Section ~\ref{Sec:Conclusion}.

\section{Related Work} \label{Sec:Related}
\subsection{Anticipation of the Future Information}
Future prediction is an emerging research area in computer vision and machine learning.
There have been many attempts to anticipate images of future to visualize dynamics of a scene in the future.
There exist the works of video prediction task, that generate the raw future frames, using the pixel distance loss function~\cite{srivastava2015unsupervised,kalchbrenner2017video}, and the generative adversarial model~\cite{vondrick2017generating,kwon2019predicting}.
Moreover, the predicted video frames are used for boosting the performance of the other video understanding tasks.
In~\cite{gao2018im2flow,rodriguez2018action}, the methods generate the future dynamic images using the current information to boost the performance of action recognition and action anticipation.
The work of~\cite{chaabane2020looking} generates the future frames with the encoder-decoder network for predicting the pedestrian's future action.

Another line of research is generating high-level representation of future frames for video understanding tasks.
In~\cite{luc2017predicting}, their semantic segmentation model directly predicts the future segmentation and shows better performance than predicting and then segmenting future RGB frames.
The work of~\cite{vondrick2016anticipating} proposes to learn visual representation of future by training unlabeled videos, and successfully predicts future action of few seconds later. 
In~\cite{wang2019early,gammulle2019predicting}, the method predicts incomplete action sequences by strengthening their features with GANs, which generates the features of unobserved videos.
Our work also stands on the representation level which can be efficiently applicable to online action detection.
Aside from exploiting the feature distance~\cite{vondrick2016anticipating} or adversarial methods~\cite{wang2019early,gammulle2019predicting}, our proposed network uses cycle-consistency to get state-of-the-art performance.

\subsection{Online Action Detection}
First defined in \cite{de2016online}, online action detection assumes that the whole length of videos is not observable at the test time, and the input to the system is streams of videos
frame-by-frame.
Work on online action detection is just scratching the surface, and only a few works have been introduced.
Due to the nature of the problem, most previous works are based on RNN structure rather than using temporal convolutions as in action recognition.

As the first paper to present the problem, ~\cite{de2016online} also introduced a dataset for online action detection called the TVSeries dataset.
The paper also provided several baseline methods based on CNN, LSTM, and SVM.
The work was extended in \cite{de2018modeling} by proposing a two-stream feedback network --  one stream for the interpretation of inputs and the other stream for modeling the temporal relation of actions.
The split streams successfully reduced the workload on classifying actions, increasing the performance of the online action detection task. 

In \cite{gao2017red}, authors tackle the action anticipation problem, a task of detecting an action before it happens.
In their work, online action detection was handled as a special case of action anticipation.
They proposed Reinforced Encoder-Decoder (RED) network, which has a reinforcement module for the sequence-level supervision.
The reinforcement module rewards the system when it makes correct predictions as early as possible.

Temporal Recurrent Network (TRN) that achieves the state-of-the-art performance in online action detection was recently introduced in \cite{xu2019temporal}.
In TRN, online action detection and anticipation of the immediate future are performed simultaneously by accumulating the historical evidences and the predicted future information.

Inspired by TRN, we also approach the online action detection problem through finding an efficient way to anticipate the future and combine it with the past information.
We were able to achieve the state-of-the-art performances on several online action detection benchmark datasets with a simpler RNN-based architecture with the following contributions:
\begin{itemize}%[topsep=2pt,leftmargin=*]
    \item 
    For anticipating futures, we apply the cycle-consistency loss in an unsupervised way rather than using the supervision from the ground truth future labels. The underlying intuition of our loss is that good future predictions should also propagate well back to the current state (Section~\ref{sec:anticipating_feature}).
    \item 
    As actions in nearby frames should be temporally consistent, we propose a method to constrain the action predictions to be temporally smooth when aggregating the past and the future information in the online setting (Section~\ref{sec:smooth_prediction}).
    \item
    When running RNN-based models, problems occur when the sequence length of videos during the inference time is much longer than that of the training. 
    We propose a novel solution to this overlooked problem to further enhance the detection accuracy
    (Section~\ref{sec:multiple_infer}).
\end{itemize}
\begin{figure*}[t]
\begin{center}
\includegraphics[width=1\linewidth]{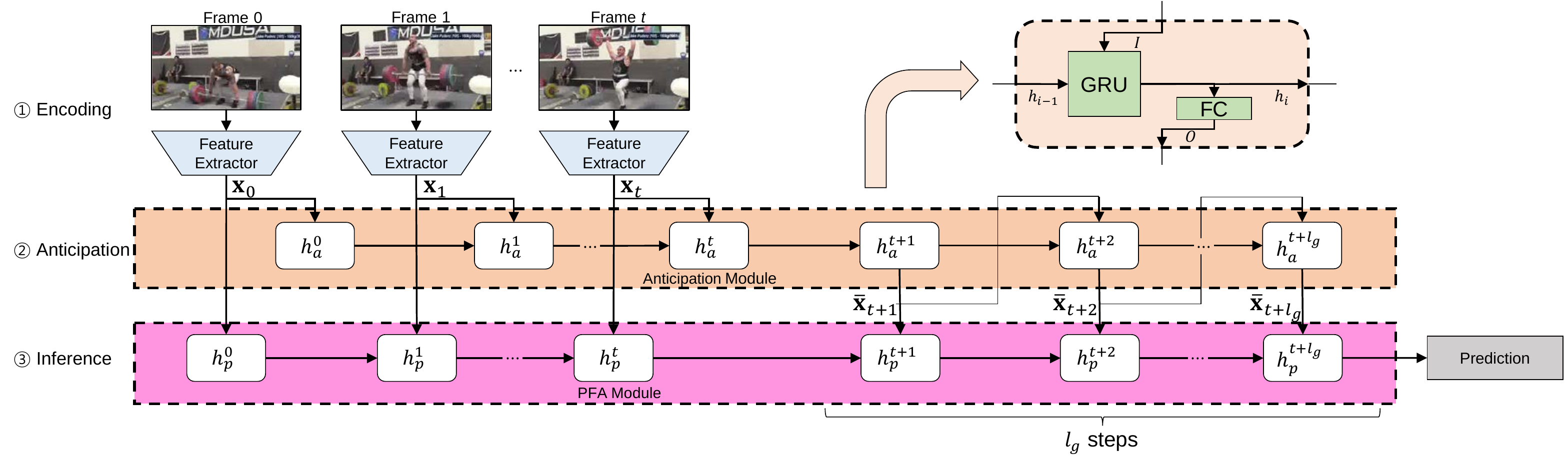}
\end{center}
\vspace{-0.5cm}
  \caption{
  Overview of our framework for predicting the action at frame $t$.
  Our method consists of a feature extractor and two RNN-based networks (anticipation module and past-future aggregation module (PFA)). Even with a very simple architecture, we achieve the state-of-the-art with several novel techniques proposed in the paper.
  }
\label{fig:overall_network}
\end{figure*}
%
%\bigskip
%
\section{Approach}  \label{Sec:Approach}
The problem of online action detection is defined as estimating the action class probabilities of each frame in videos under the online setting.
By the property of the online setting, only past video frames $\mathbf{V}=\{\mathbf{I}_0,\mathbf{I}_1,\cdots,\mathbf{I}_t\}$ are observable at timestamp $t$, and further frames $\{\mathbf{I}_{t+1},\mathbf{I}_{t+2},\cdots\}$ are not observable.
Given $\mathbf{V}$, our task is to predict the probabilities $\mathbf{p_t}=\{p_t^i\}_{i=0}^{K}$ of K+1 action classes, including the background class, continuously with increasing $t$.

\subsection{Algorithm Overview}
As shown in Fig.~\ref{fig:overall_network}, our framework  goes through three stages -- feature encoding, future anticipation, and inference by combining the past and the future information -- to solve the online video detection problem. 
Network components include the feature extractor, the anticipation module, and the past-future aggregation (PFA) module. 

In the first encoding stage, the feature extractor computes the compact representation $\mathbf{x}_t$ of the input frame $\mathbf{I}_t\in\mathbb{R}^{H\times W}$.
In our work, we use the two-stream network in \cite{xiong2016cuhk} pretrained on ActivityNet~\cite{caba2015activitynet} for the feature extractor that generates 1-D feature vector $\mathbf{x}_t$. The size of the vector is 3072\footnote{Note that we also use Kinetics pretrained network with the vector size of 2048 in the experiments.}.
The stream of $1$-D vectors $\mathbf{x}=\{\mathbf{x}_0,\mathbf{x}_1,\cdots,\mathbf{x}_t\}$ is processed by the rest of our networks. 

In the anticipation stage, we use the anticipation module to generate features of the future $\{\bar{\mathbf{x}}_{t+1},\bar{\mathbf{x}}_{t+2},\cdots,\bar{\mathbf{x}}_{t+{l_g}}\}$.
Anticipation module is an RNN-based network, which encodes the input features to its hidden state and generates the anticipated feature of the next frame.

The last stage is the inference stage that outputs the action prediction at the current frame using the PFA module. 
PFA module is another RNN-based network, which keeps encoding both inputs (past) and generated features (future) to its hidden state.
PFA module also contains a classifier with fully-connected layers, which produces the probabilities of action classes after its last feature encoding.

As one can see, we use a very simple architecture that employs a feature extractor and two RNNs.
While our framework shares the same strategy of combining the current and the future information through an anticipation method as in TRN~\cite{xu2019temporal}, the implementation of the strategy is quite different.
In TRN, anticipated features at every future time step are average-pooled together and transformed via a FC layer to generate the final future representation. This is then combined with the current features to construct the input for the RNN that makes the prediction.
In contrast, we assume that the features from the anticipation module (e.g. $\bar{\mathbf{x}}_{t+1}$) can replace the original input features (e.g. $\mathbf{x}_{t+1}$) and employ a simple 2-layer RNN system. 
Even with our simple structure, we achieve the state-of-the-art performances by using several concepts that are explained in the following subsections. 

\subsection{Anticipating Features with Cycle-consistency} \label{sec:anticipating_feature}
Different from the offline setting, we cannot use the information from the future frames in the online setting.
This lack of temporal information causes the performance gap between offline action detection and online action detection.
Filling in the future information on its own, \ie future anticipation, is one of the solutions that can close this gap.

One way to train the anticipation task is through supervision by exploiting future frame information (e.g. class labels in the future) as ground-truth since it is provided in the training set.
This strategy was used in TRN~\cite{xu2019temporal}.
In TRN, the anticipation module was trained by the cross-entropy loss using the ground truth class labels in the future frames. 
\begin{figure*}[t]
\begin{subfigure}{0.49\textwidth}
\includegraphics[width=1\linewidth]{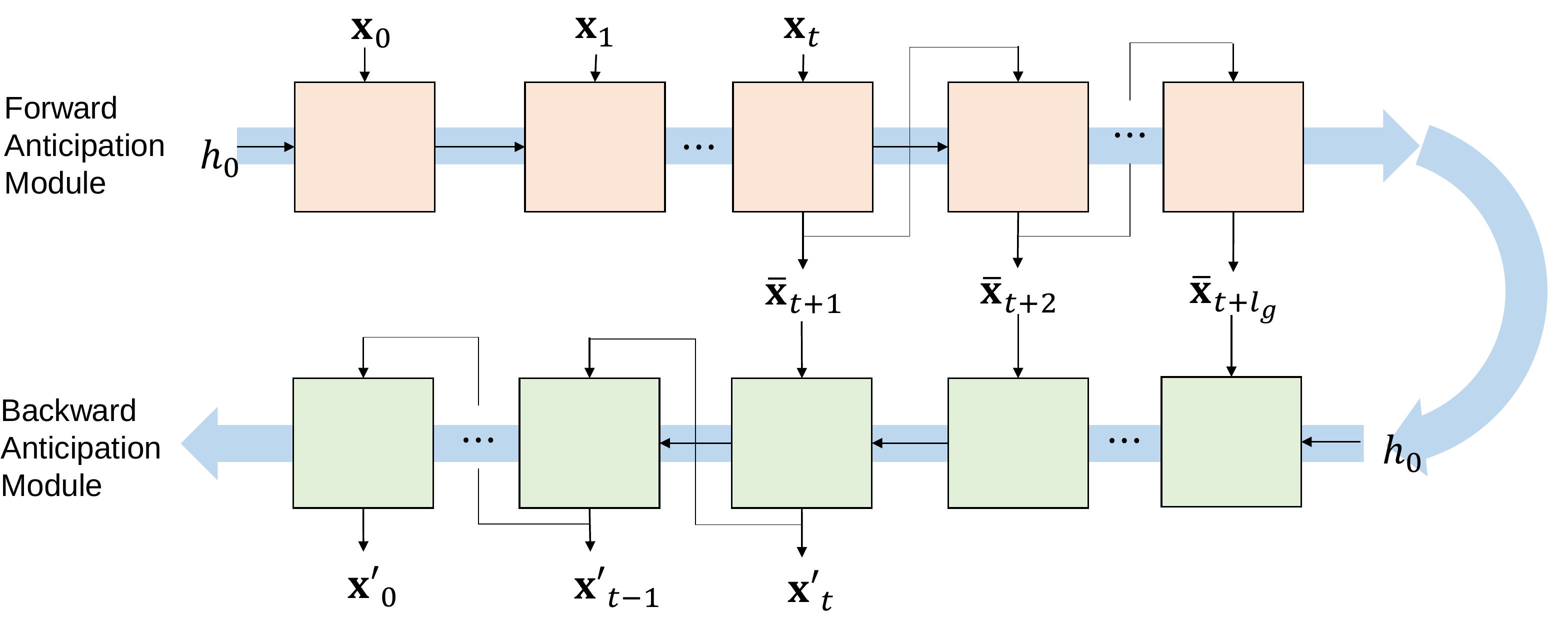}
\caption{The first phase training of the anticipation module.}
\label{fig:anticipating_feature_1}
\end{subfigure}
\begin{subfigure}{0.49\textwidth}
\includegraphics[width=1\linewidth]{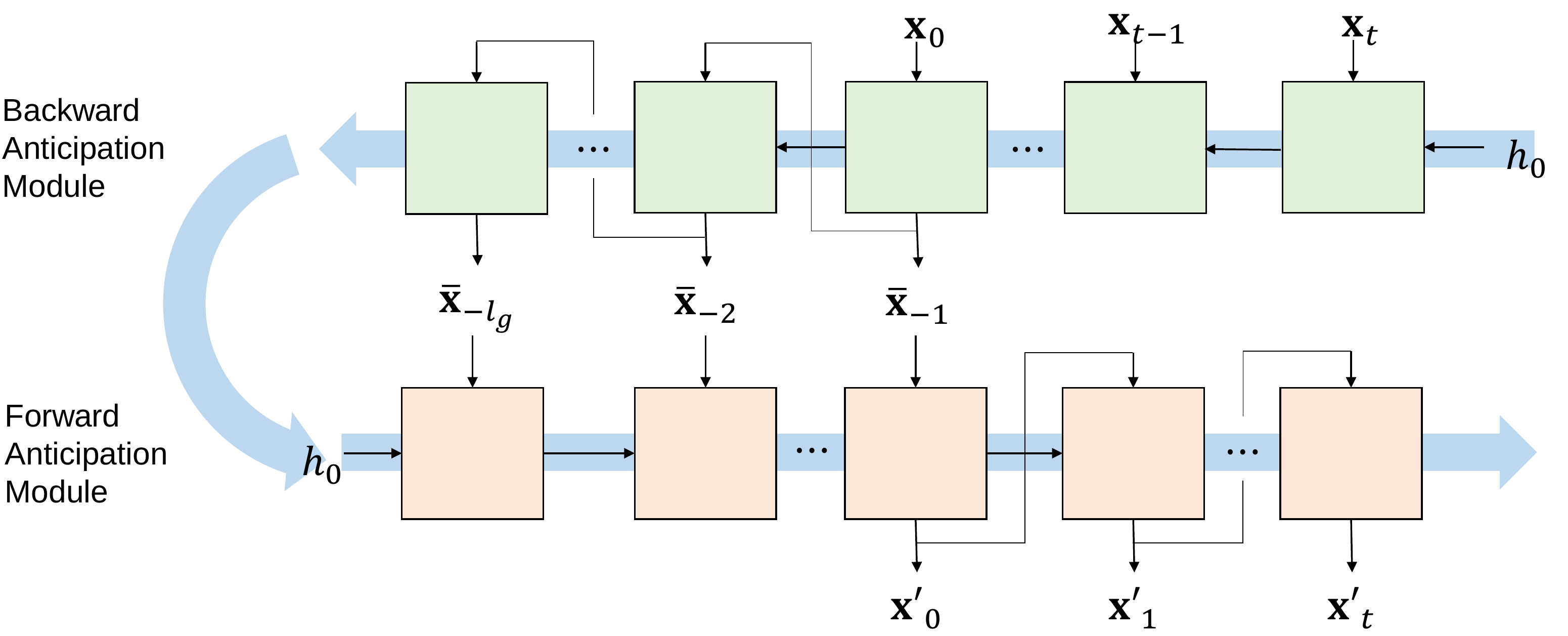}
\caption{The second phase training of the anticipation module.}
\label{fig:anticipating_feature_2}
\end{subfigure}
  \caption{Two-phase training of the anticipation module. 
  Note that the fully-connected decoder is shared in the forward anticipation module and the backward anticipation module.
  L2 loss is computed from the $t$ input features and the generated $t$ features from the counterpart anticipation module.}
\label{fig:anticipating_feature}
\end{figure*}

In this work, we take a different direction for solving the anticipation problem and adopt unsupervised learning.
Recently, cycle-consistency has shown great success in the image-to-image translation problem \cite{zhu2017unpaired} and has been adopted in many following works. 
In a way, the feature anticipation task in videos is similar to the image-to-image translation as we target at translating past frames to future frames on the feature level. 
Therefore, we exploit cycle-consistency in our work to train the anticipation module in an unsupervised manner. 

For the past set $X$ and the future set $Y$, our goal is to learn an anticipation function $G:X\rightarrow{Y}$.
To constrain the training of $G$, we also set a past generation function $F:Y\rightarrow{X}$ and enforce $F(G(X))\approx X$.
With this constraint, we do not need ground truth labels of the future frames to train the anticipation module, allowing our work to generalize better with less overfitting.

As for the details of our training method, we adopt two GRU cells as the forward and the backward feature encoders, and one fully-connected layer as the shared feature decoder.
We call them as $\mathrm{fGRU}$, $\mathrm{bGRU}$, and $\mathrm{dec}$, respectively.
The forward pass generates the feature of the next frame and the backward pass generates the feature of the previous frame.
In the test time, $\mathrm{bGRU}$ is discarded, and $\mathrm{fGRU}$ and $\mathrm{dec}$ are used as the anticipation module.

In the first training phase (Fig.~\ref{fig:anticipating_feature_1}), $\mathrm{fGRU}$ encodes from 0 to $t$-th frame features starting with all-zero hidden state $h_0$,
\begin{equation}
h_{t+1}=\mathrm{fGRU}(\mathbf{x}[0:t], h_0).
\end{equation}
Then, $\mathrm{dec}$ generates the future frame feature $\mathbf{\bar{x}}_{t+1}$ and $\mathrm{fGRU}$ continues the encoding with the generated $\mathbf{\bar{x}}_{t+1}$:
\begin{align}
\mathbf{\bar{x}}_{t+1} &=\mathrm{dec}(h_{t+1}), \\
h_{t+2} &=\mathrm{fGRU}(\mathbf{\bar{x}}_{t+1}, h_{t+1}).
\end{align}
This process is repeated for $l_g$ times, and we gain $\mathbf{\bar{x}}=\{\mathbf{\bar{x}}_{t+1},\mathbf{\bar{x}}_{t+2},\cdots,\mathbf{\bar{x}}_{t+l_g}\}$.
For the backward pass, $\mathrm{bGRU}$ takes $\mathbf{\bar{x}}$ as input, and $\mathrm{dec}$ generates features $\mathbf{x}'=\{\mathbf{x}_0',\mathbf{x}_1',\cdots,\mathbf{x}_t'\}$.
We calculate the cycle consistency as the L2 loss between the input $\mathbf{x}$ and the generated feature $\mathbf{x}'$,
\begin{equation}
L_{cc}=\sum_{i=0}^{t} (\mathbf{x}_i'-\mathbf{x}_i)^2.
\end{equation}

In the first training phase, training $\mathrm{bGRU}$ can be difficult as $\mathrm{bGRU}$ can only observe the generated features as the input.
To avoid this issue, we invert all the inputs and the network flows in the second training phase to enable training $\mathrm{bGRU}$ with real features, not just the generated features. 
This is shown in Fig.~\ref{fig:anticipating_feature_2}.
$\mathrm{bGRU}$ takes $\mathbf{rx}=\{\mathbf{x}_t,\mathbf{x}_{t-1},\cdots,\mathbf{x}_0\}$ as input, and $\mathrm{dec}$ generates $\mathbf{\overline{rx}}=\{\mathbf{\bar{x}}_{-1},\mathbf{\bar{x}}_{-2},\cdots,\mathbf{\bar{x}}_{-l_g}\}$.
Then, forward encoder takes $\mathbf{\overline{rx}}$, generates $\mathbf{rx}'$ again, and calculate the L2 Loss between $\mathbf{rx}$ and $\mathbf{rx}'$.

\subsection{Past and Future Aggregation with Temporally Smooth Prediction}\label{sec:smooth_prediction}
\begin{figure}[t]
\begin{minipage}[b]{0.49\linewidth}
 \centering
 \includegraphics[width=1\linewidth]{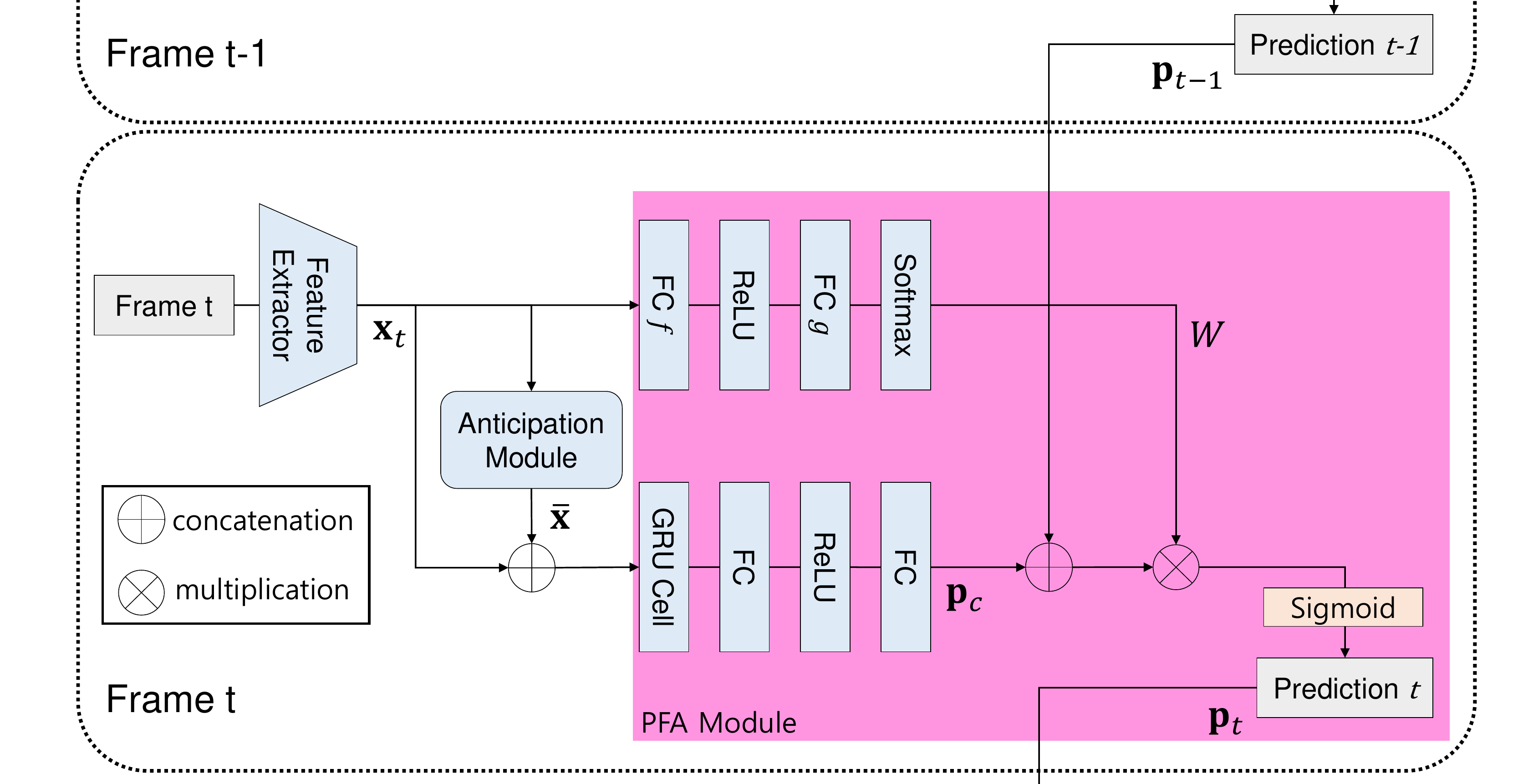}
  
\end{minipage}
\hfill
\begin{minipage}[b]{0.49\linewidth}
 \centering
 \includegraphics[width=1\linewidth]{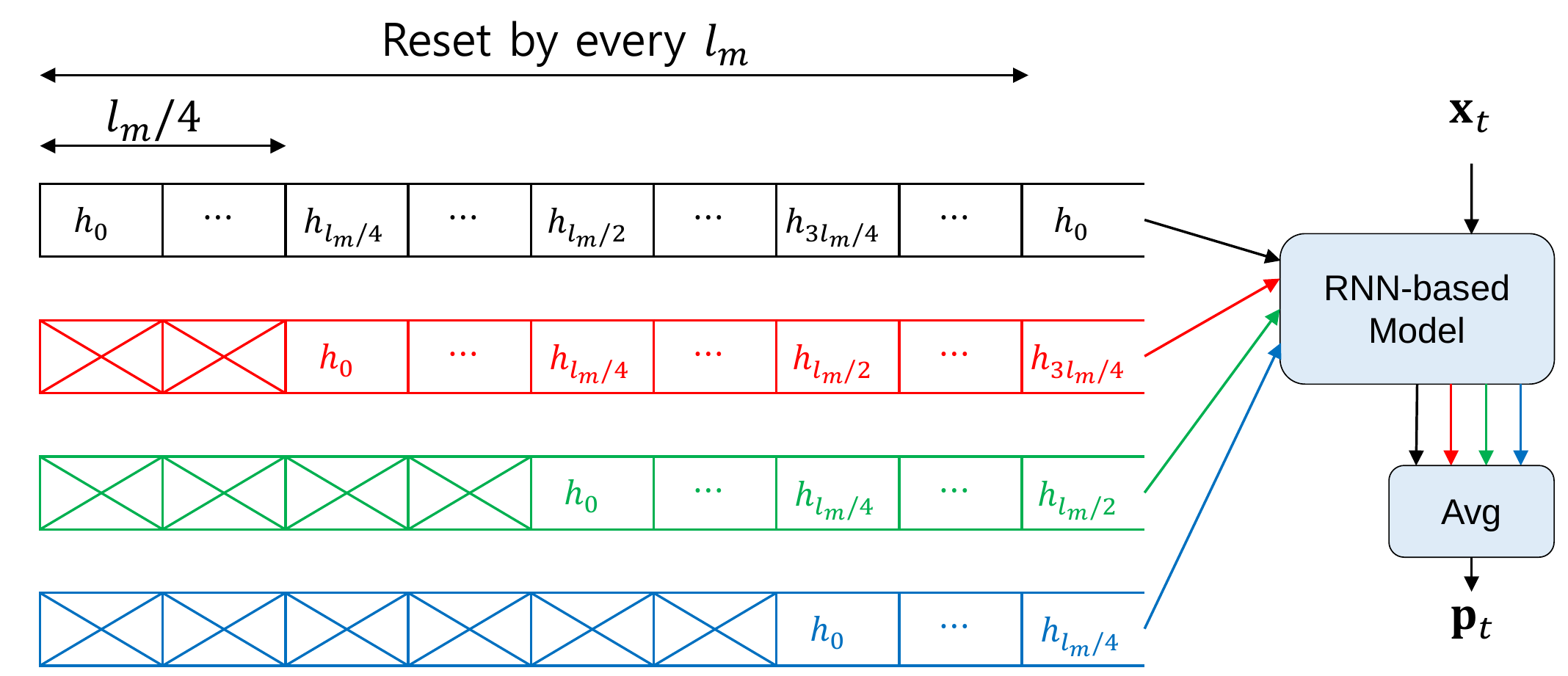}
\end{minipage}
\caption{%
    \textit{(left)}
    Past-future aggregation (PFA) module at frame $t$ for temporally smooth predictions. 
    Weight parameter $\mathbf{w}$ is generated from the weight generator (4-layer network on the top). 
    The classifier (4-layer network on the bottom) makes the action probability $\mathbf{p}_{c
    }$.
    Final prediction is made by concatenating $\mathbf{p}_c$ and $\mathbf{p}_{t-1}$ with the computed weight $\mathbf{w}$.
    \textit{(right)}
    Multiple inference during the test time.
    We run $n$ independent inferences ($n=4$) with the interval of $l_m/4$ during the test time to deal with long video sequences. 
    Each hidden state is reset after reaching the training length $l_m$.
    All inferences are averaged to make the final prediction $p_t$.
    This technique can be applied to any RNN-based models, to deal with long sequences during the test time.
 }
 \label{fig:sp_and_mi}
\end{figure}

The last phase of the training is to train the past-future aggregation (PFA) module (Fig.~\ref{fig:sp_and_mi}).
The input to PFA module is the past input features and the generated future features.
After processing the input through several layers (GRU-FC-ReLU-FC), K+1 action probabilities $\mathbf{p_c}=\{p_c^i\}_{i=0}^{K}$ are computed.

To further improve the classification performance of the PFA module, we additionally propose a novel method to enforce the temporal smoothness in between the predictions of nearby frames.
It is reasonable to assume that action classes between adjacent frames will be the same in most cases, therefore, the action probabilities should not change drastically on nearby frames.

Traditionally, window-based temporal smoothing techniques such as non-maximum suppression (NMS) have been used in video tasks~\cite{THUMOS14} as a post-processing at inference time.
However, it is difficult to apply it to our task since we can only access past frames in the online setting.
Using the past information, we propose a temporally coherent structure of RNN to compute the final classification probabilities at frame $t$ ($\mathbf{p}_t$) by considering the current prediction ($\mathbf{p}_c$) as well as the prediction in the previous frame ($\mathbf{p}_{t-1}$).

We add two fully-connected layers $f$ and $g$ on the input feature $\mathbf{x}_i$ to generate one-dimensional vector $\mathbf{w}$ of size 2: 
\begin{equation}
\mathbf{w}=\mathrm{softmax}(g(\mathrm{ReLU}(f(\mathbf{x}_i)).
\end{equation}
$\mathbf{w}$ balances the weight between the current probabilities $\mathbf{p}_c$ and the previous probabilities $\mathbf{p}_{t-1}$.
The final probability at frame $t$ ($\mathbf{p}_t$) can be computed as follows:
\begin{equation}
    \mathbf{p}_t=\mathbf{w}\times
    \begin{bmatrix}
        \mathbf{p}_{c} \\
        \mathbf{p}_{t-1}
    \end{bmatrix} .
\end{equation}

We optimize PFA with the cross-entropy loss using the ground-truth label $\mathbf{y_t}=\{y_t^i\}_{i=0}^{K}$, 
\begin{equation}
L_{PFA}=-\sum_{i=0}^{K} \frac{1}{K+1}[y_t^i\cdot \log p_t^i +(1-y_t^i)\cdot \log (1-p_t^i)].
\end{equation}

Alg.~\ref{alg:training} details the training procedure for our whole network. 
\begin{algorithm}[ht]
\caption{Training procedure}\label{alg:training}
\begin{algorithmic}[1]
\State Initialize the parameters of the network
\While{training not converge}
    \State $\mathbf{x}\gets$ Obtain a training sequence of length $l_m$
    \For{$t=1:l_m$}
        \State $\mathbf{\bar{x}}\gets$ ForwardPass($\mathbf{x}[0:t],l_g$)
        \State $\mathbf{x}'\gets$ BackwardPass($\mathbf{\bar{x}},t$)
        \State Calculate $L_{cc}$ using ($\mathbf{x}[0:t],\mathbf{x}'$) 
        \State Update parameters of $\mathrm{fGRU}, \mathrm{bGRU}, \mathrm{dec}$
        \State
        \State $\mathbf{\overline{rx}}\gets$ BackwardPass($\mathbf{x}[t:0],l_g$)
        \State $\mathbf{rx}'\gets$ ForwardPass($\mathbf{\overline{rx}},t$)
        \State Calculate $L_{cc}$ using ($\mathbf{x}[t:0],\mathbf{rx}'$)
        \State Update parameters of $\mathrm{fGRU}, \mathrm{bGRU}, \mathrm{dec}$
        \State
        \State $feature\gets$ Merge $\mathbf{x}$ and $\mathbf{\bar{x}}$ 
        \State $\mathbf{p}_t\gets$PFA($feature$)
        \State Compute $L_{PFA}$ by $\mathbf{p}_t$
        \State Update classifier
    \EndFor
\EndWhile
\end{algorithmic}
\end{algorithm}

\subsection{Multiple Inference during Test Time}\label{sec:multiple_infer}
When continuously detecting actions on a long streamed video during the test time, it is important to accumulate the past information properly as the performance of RNN-based models is severely affected by the error accumulation.
Usually, a long sequence is processed by either simply accumulating the whole sequence or splitting the sequence into non-overlapping subsequences.
However, the first approach experiences serious performance loss due to the error accumulation.
In addition, saving a long-term past may be inefficient in our task as the duration of an action is usually not too long.
The second approach may be reasonable in our task as our model works well when the length of the subsequences is the same as the one used for the training. 
In this case, however, temporal information near the boundary of subsequences is lost, resulting in a performance loss.

To tackle this issue, we propose a simple yet effective approach that can be applied to any RNN-based models.
As shown in Fig.~\ref{fig:sp_and_mi},
we run $n$ independent inferences using the trained model, each with independent hidden states. 
Each inference starts with the interval of \( \frac{l_m}{n} \), and is reset after the training length $l_m$.
$n$ inferences are averaged to generate the final output.

Our method can relieve the two aforementioned negative issues on RNN-based models:
\begin{enumerate*}[label={\arabic*)}]
	\item By resetting the hidden state, the model can process proper length of inputs, \ie, the length of the trained inputs. 
	\item By exploiting $n$ hidden states with different frame interval, the final prediction for the frame classification takes advantage of the multi-scale temporal information. 
\end{enumerate*}

\section{Experiments}\label{sec:experiments}
\subsection{Datasets}
\noindent\textbf{TVSeries} dataset was introduced in~\cite{de2016online} while defining the online action detection task.
It consist of 27 episodes of 6 popular TV series with temporal annotations at the frame-level for 30 action classes.
The total length of videos spans over 16 hours and it has a total of 6,231 action instances. 

\noindent\textbf{THUMOS'14} can be applied for various video understanding tasks including temporal action localization~\cite{THUMOS14}. 
Among 101 action classes included in the dataset, 20 action classes are available for the temporal action localization task.
The temporal annotations are available on the validation set (200 untrimmed videos) and the test set (213 untrimmed videos), with 6k action instances.
We used the validation set for the training, and the test set for the evaluation.

\noindent\textbf{BBDB} is a baseball specific dataset presented in~\cite{shim2018teaching} and it can be also used for various video understanding tasks including temporal localization.
It contains 1,172 full baseball game videos with a total of 4,254 hours of untrimmed video.
It also contains over 400k temporal annotations over 30 baseball activities. 
We used its localization setup and applied it to the per-frame online action detection task.

\subsection{Experimental Settings}
\noindent\textbf{Evaluation Metrics}. 
We follow the evaluation metrics of each dataset from the existing works. 
We use per-frame mean average precision (mAP) on THUMOS'14 and BBDB.
On TVSeries, we use per-frame calibrated average precision (cAP) as proposed in \cite{de2016online}.
cAP can relieve the imbalance of the positive and the negative samples by calculating the calibrated precision as
\begin{equation}
cPrec=\frac{TP}{TP+\frac{FP}{w}},
\end{equation}
where $w$ is the ratio between the negative frames and the positive frames. 
cAP can be calculated as the same way as mAP using $cPrec$, \(cAP=\sum_{k}cPrec(k)*I(k)/P\), where $P$ is the total number of positive frames, and $I(k)$ is equal to 1 if frame $k$ is a true positive and equal to 0 otherwise.

\noindent\textbf{Implementation Details}.
For the evaluations on TVSeries and THUMOS'14, we test two versions of our work with different feature extractors.
On one version (FATSnet-ActivityNet), we used the two-stream model in~\cite{xiong2016cuhk} that is trained on ActivityNet~\cite{caba2015activitynet}. The features from this model were also used in other previous works.
On the other version (FATSnet-Kinetics), two-stream TSN~\cite{wang2016temporal} trained on Kinetics~\cite{carreira2017quo} was used for the feature extractor. 
Since Kinetics is one of the largest datasets for video understanding tasks, there exists a trend of exploiting this rich information in other video understanding studies.
As the input feature of our main network, we used the output of the last flatten layer of each model.
For frame sampling, we set a video of 0.25 seconds as a chunk, and the overall process flows by the chunk level. 

On BBDB, we trained RGB-stream TSN~\cite{wang2016temporal} with ResNet-50~\cite{he2016deep} backbone network on the action recognition settings of BBDB, and used it as the feature extractor.
We extract the video frames at 6 frames per second, and take the output of the last flatten layer as the input feature for our main network.

For optimization, we used Adam~\cite{kingma2014adam} optimizer with the learning rate of 0.00005, and the batch size of 32.
We set the maximum length of the input sequence $l_m$ as 32, and the default inference steps at test time as 4.

\begin{table}[t]
\begin{center}
\begin{tabular}{|p{0.35\linewidth}p{0.075\linewidth}|p{0.35\linewidth}p{0.075\linewidth}|}
\hline
\multicolumn{2}{|c|}{THUMOS'14} & \multicolumn{2}{|c|}{TVSeries} \\
\hline
Method     & mAP    & Method     & mcAP\\ \hline \hline
Single-frame CNN~\cite{simonyan2014very}      & 34.7  & CNN~\cite{de2016online}      & 60.8       \\
Two-stream CNN~\cite{simonyan2014two}      & 36.2   & LSTM~\cite{de2016online}      & 64.1       \\
C3D+LinearInterp~\cite{shou2017cdc}     & 37.0  & RED-VGG~\cite{gao2017red} & 71.2 \\
LSTM~\cite{yeung2018every}          & 39.3    & Stacked LSTM~\cite{de2018modeling} & 71.4  \\
MultiLSTM~\cite{yeung2018every} & 41.3      & 2S-FN~\cite{de2018modeling}    & 72.4  \\
Conv\&De-Conv~\cite{shou2017cdc}    & 41.7  & FV + SVM~\cite{de2016online} & 74.3              \\
CDC~\cite{shou2017cdc} & 44.4        & TRN-VGG~\cite{xu2019temporal}  & 75.4   \\
RED~\cite{gao2017red} & 45.3        & RED-TS~\cite{gao2017red} & 79.2  \\
TRN-TS~\cite{xu2019temporal}  & 47.2    & TRN-TS~\cite{xu2019temporal}  & \textbf{83.7}               \\ \hline
FATSnet-ActivityNet & \textbf{51.6}      & FATSnet-ActivityNet & 81.7       \\ 
FATSnet-Kinetics & \textbf{59.0}     & FATSnet-Kinetics & \textbf{84.6}      \\ \hline
\multicolumn{2}{|c|}{BBDB} & & \\
\hline
Method     & mAP    & &\\ \hline\hline
Single frame ~\cite{shim2018teaching}      & 9.25  & &\\
CDC~\cite{shou2017cdc} & 23.3   & &        \\
TRN~\cite{xu2019temporal}  & 27.6       & &      \\ \hline
FATSnet  & \textbf{29.0} & &     \\ \hline
\end{tabular}
\end{center}
\vspace{-0.6cm}
\caption{Quantitative results on benchmark datasets. 
All numbers on the tables are as reported in other papers except for the results on BBDB. 
}
\label{tab:quantitative}
\end{table}

\subsection{Quantitative results}
Quantitative results on benchmark datasets are shown in Table~\ref{tab:quantitative}.
On THUMOS'14, both versions of our method outperform all previous work by a good margin. 
When using the pretrained features on Kinetics (FATSnet-Kinetics), the performance skyrockets to mAP of 59\%, an increase of nearly 12\% over the current state-of-the-art measure from TRN that uses two-stream features.
On TVSeries dataset, FATSnet-Kinetics outperforms TRN by about 1\% and FATSnet-ActivityNet shows 2\% less accuracy compared to TRN. 

Note that the differences between the two datasets are the density of action instances and the diversity of scenes.
THUMOS'14 includes high density of action instances, while TVSeries contains data with low density of action instances and large diversity of scenes.
We have shown that our proposed method can deal with both cases very well in general and can also benefit from better input features. 

On BBDB, we only compared with a few methods as not many algorithms have been tested on BBDB.
We used the same input setting (RGB-stream feature) on each method.
Overall accuracy numbers are much lower on BBDB compared to other datasets, due to similarities between actions in baseball. 
On BBDB, our method performs slightly better the TRN. 

Our model's inference speed is 9.9 fps which is faster than the sampling speed of input videos (4 fps on THUMOS and TVSeries, and 6 fps on BBDB), which means that our model is practical enough to be used in real-time.
\begin{figure*}[t]
\begin{subfigure}{\textwidth}
\includegraphics[width=1\linewidth]{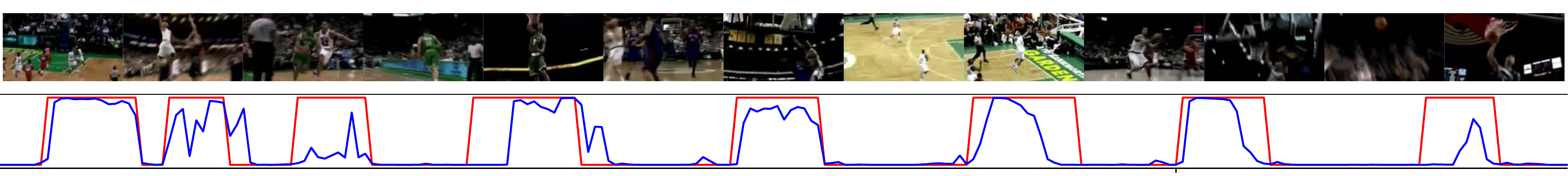}
\caption{Qualitative results on \textit{BasketballDunk} of THUMOS'14 showing confidence score frame by frame.}
\label{fig:qualitative_sub_thumos}
\end{subfigure}
\begin{subfigure}{\textwidth}
\includegraphics[width=1\linewidth]{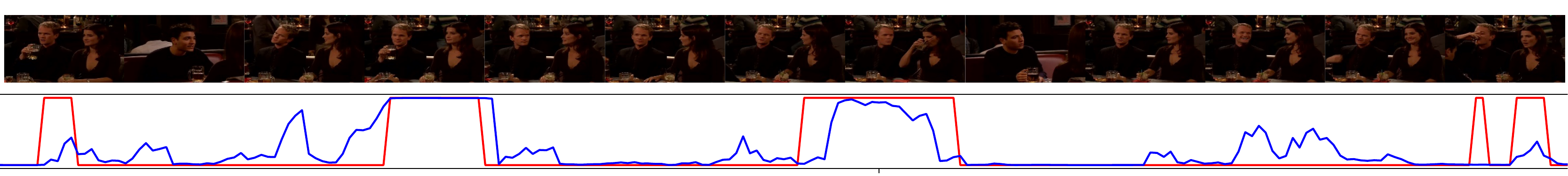}
\caption{Qualitative results on \textit{Drink} of TVSeries.}
\label{fig:qualitative_sub_tvseries}
\end{subfigure}
\begin{subfigure}{\textwidth}
\includegraphics[width=1\linewidth]{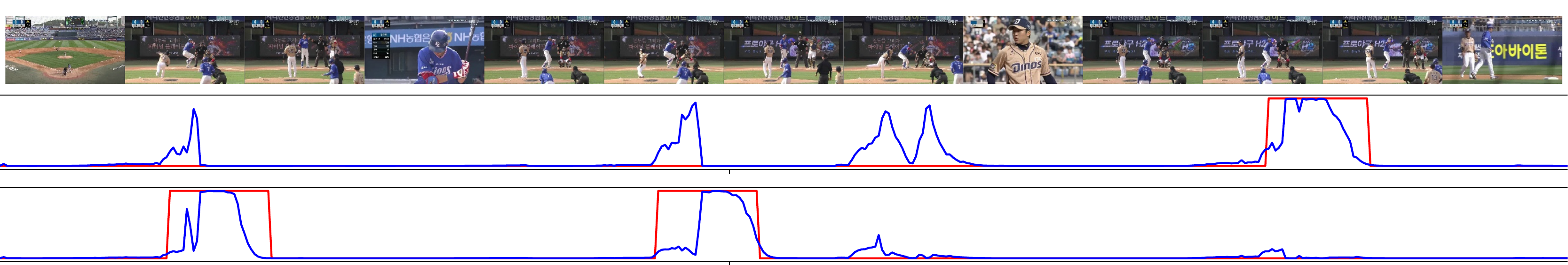}
\caption{Qualitative results on \textit{Ball} and \textit{Strike} of BBDB.
The second row is the result of \textit{Ball} class, and the the third row is the result of \textit{Strike} class.}
\label{fig:qualitative_sub_bbdb}
\end{subfigure}
  \caption{Qualitative results. Best viewed in color.
  \textcolor{red}{Red lines} are the ground truth and \textcolor{blue}{blue lines} are the confidence score of our network.}
\label{fig:qualitative_result}
\end{figure*}

\subsection{Qualitative results}
As a qualitative result, we first present a visualization of confidence scores by frames on THUMOS'14 (Fig.~\ref{fig:qualitative_sub_thumos}) and TVSeries (Fig.~\ref{fig:qualitative_sub_tvseries}).
FATSnet generally makes accurate prediction on each frame, and shows decent boundaries of action instances.

Fig.~\ref{fig:qualitative_sub_bbdb} shows the results of \textit{Ball} and \textit{Strike} class on BBDB.
Note that discriminating \textit{Ball} and \textit{Strike} class is very challenging; the only difference is the position where the ball is caught.
FATSnet predicts the sequence as \textit{Ball} first, and changes its decision as \textit{Strike} after it observe the position of the ball.

\subsection{Ablation Study}
\subsubsection{Effectiveness of Each Component}
We first examine the results by changing the combinations of the modules proposed in our work.

As can be seen in Table~\ref{tab:eff_contrib2}, each proposed contribution shows notable increase in the performance regardless of datasets. 
We also test the influence of changing $l_g$ (Table~\ref{tab:eff_contrib}), which is the length of the future generation. 
The performance goes up as $l_g$ is increased, and the maximum value is reached at $l_g=8$ or $l_g=12$.

\begin{figure}[t]
\begin{center}
\includegraphics[width=0.85\linewidth]{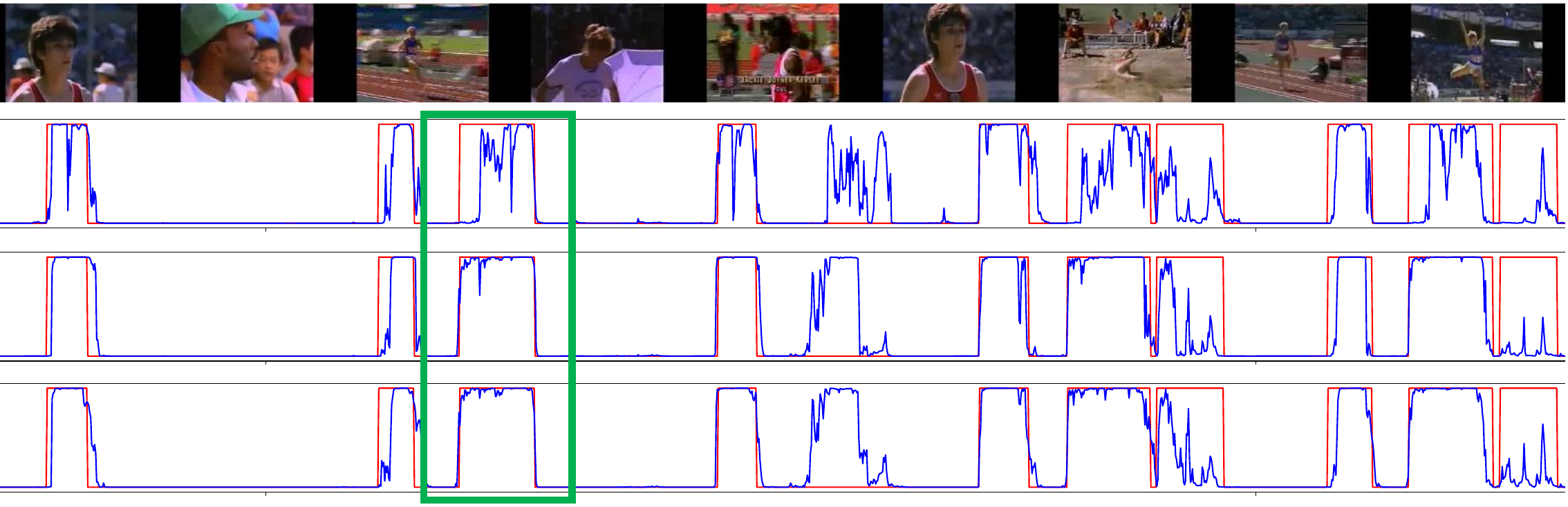}
\end{center}
  \caption{Results on \textit{LongJump} of THUMOS'14 showing confidence score frame by frame.
The second row is the result of ATCP, and the the third row is the result ATCP+SP.
The last row is the result of our final model.}
\label{fig:ablation_each_step}
\end{figure}

We also show a visualization of confidence scores by frames on \textit{LongJump} sequence in THUMOS'14 (Fig.~\ref{fig:ablation_each_step}).
The highlighted box in Fig.~\ref{fig:ablation_each_step} is a great example for an explanation. 
The confidence scores without the temporal smoothing and the multiple inference fluctuate in most parts of the data.
With the temporal smoothing, we can observe the drop in fluctuations and overall smooth scores.
At this point, it is also observed that the confidence score significantly drops by the reset of the hidden states.
Our final model that also includes the multiple inference reduces all the issues and shows decent boundaries of action instances.

\begin{table}[t]
\begin{center}
\begin{tabular}{l|l|l|l}
\toprule
Dataset                    & Pre-trained Feature          & Approach   & mAP \\ \hline
\multirow{6}{*}{THUMOS'14}  & \multirow{3}{*}{ActivityNet} & ATCP       & 48.6    \\
                           &                              & ATCP+SP    & 49.6     \\
                           &                              & ATCP+SP+MI & 51.6     \\ \cline{2-4} 
                           & \multirow{3}{*}{Kinetics}    & ATCP       & 54.8    \\
                           &                              & ATCP+SP    & 56.8     \\
                           &                              & ATCP+SP+MI & 59.0     \\ \hline
\multirow{6}{*}{TVSeries}  & \multirow{3}{*}{ActivityNet} & ATCP       & 79.6    \\
                           &                              & ATCP+SP    & 80.9     \\
                           &                              & ATCP+SP+MI & 81.7     \\ \cline{2-4} 
                           & \multirow{3}{*}{Kinetics}    & ATCP       & 82.2    \\
                           &                              & ATCP+SP    & 83.4     \\
                           &                              & ATCP+SP+MI & 84.6    \\ \hline
\multirow{3}{*}{BBDB}      & \multirow{3}{*}{RGB}         & ATCP       & 24.5    \\
                           &                              & ATCP+SP    & 25.7    \\
                           &                              & ATCP+SP+MI & 29.0    \\ \bottomrule
\end{tabular}
\end{center}
\caption{mAPs by changing datasets and the combinations of the modules. Our approach consists of feature anticipation(ATCP), smooth prediction(SP), and multiple inference at test time(MI).
}
\label{tab:eff_contrib2}
\end{table}

\begin{table}[t]
\begin{center}
\begin{tabular}{p{0.3\linewidth} | c P{0.07\linewidth} P{0.07\linewidth} P{0.07\linewidth} P{0.07\linewidth} P{0.07\linewidth} P{0.07\linewidth}}
\toprule
Approach & $l_g$  &  0 & 4 & 8 & 12 & 16 \\\midrule \midrule
ATCP &  
    & 47.4  & 47.7  & 48.0  & \textbf{48.6}  & 48.1    \\ %\midrule
ATCP + SP &   
    & 47.8  & 48.7  & \textbf{49.6}  & 49.6  & 48.6    \\ %\midrule
ATCP + SP + MI &      
    & 49.6  & 51.2  & \textbf{51.6}  & 50.3  & 49.4    \\ \bottomrule
\end{tabular}
\end{center}
\caption{mAPs on THUMOS'14 with different combinations of steps $l_g$ in our framework. When $l_g$ is 0, future anticipation process is occluded, and predicts with past frames only.}
\label{tab:eff_contrib}
\end{table}

\subsubsection{Effectiveness of Cycle-consistency}
One of the key contributions of our work is the use of the unsupervised cycle-consistency loss for the anticipation. 
To show the effectiveness of the cycle-consistency, we compare our method with the class label supervision of the future frames used in~\cite{xu2019temporal}.
To this end, we modified our method for the class supervision.
Specifically, we added a fully-connected layer which outputs class-level predictions after the anticipation module.
We used a cross-entropy loss with the ground-truth class labels of each frame for training.
As a result, the performance of the class supervision was mAP of 47.2\%, which is lower than our method with cycle-consistency (48.6\%).
The mix model of class supervision and cycle-consistency shows the performance of 47.2\%, which does not show the performance gain over the cycle-consistency model.
Besides the class supervision, we also modified our method for the feature-level supervision.
All settings were the same as the class supervision, except using L2 loss with the ground-truth feature for the training.
The performance of the feature-level supervision was mAP of 48.2\%, which is still lower than our method. 
We conjecture that the performance gain of our method comes from the flexibility and the expressiveness in predicting future by learning in a feature space instead of using either a label space or fixed ground-truth feature space.
As action is composed of temporally varying low-level phases of action such as running, jumping, and arm swing in volleyball spike, learning such information in a feature space using our cycle-consistency is beneficial compared to class label supervision.
Also, our cycle-consistency can train flexible features of undecided future, while hard-supervision can suffer from inaccurate predictions due to the difference between the train and the test sets.

\subsubsection{Comparison for Temporal Smoothing}
\begin{figure}[t]
\begin{center}
\includegraphics[width=0.65\linewidth]{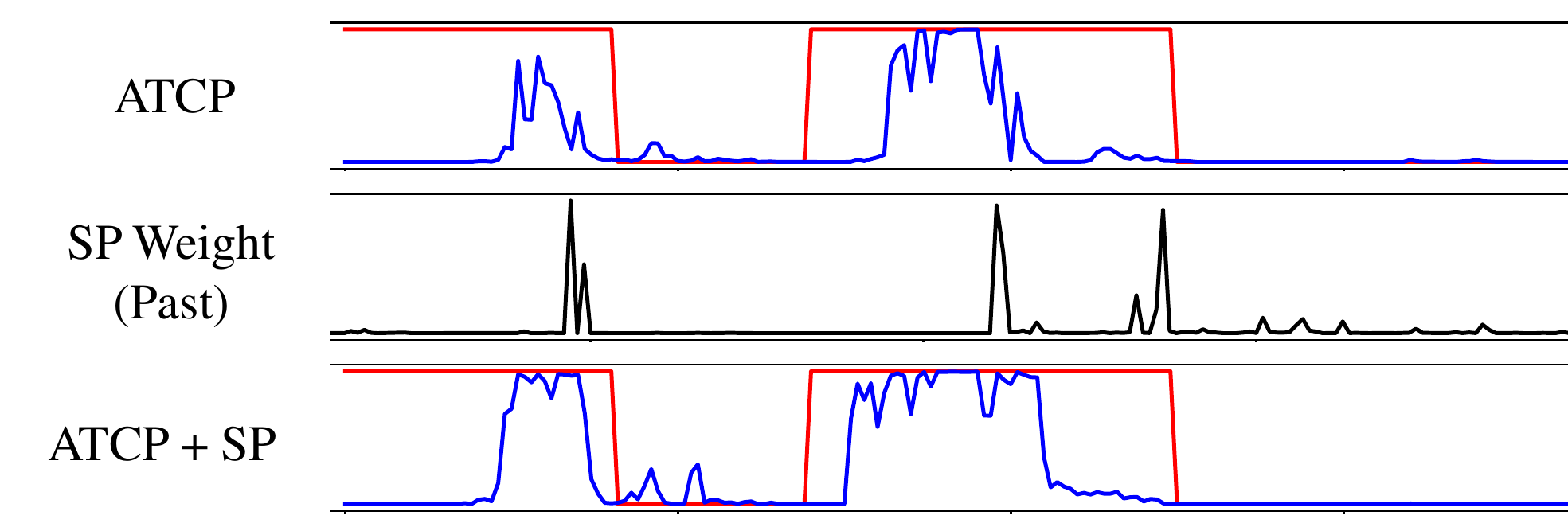}
\end{center}
  \caption{The example how the weights learned on temporal smoothing module. 
  The first row is the result of ATCP, and the the second row is the past-related learned weights of PFA module.
  The last row is the result of ATCP+SP, which aggregates the learned weights to ATCP.}
\label{fig:ablation_smoothweight}
\end{figure}

To show the effectiveness of our temporal smoothing, we compare our method with the NMS method.
Although the NMS is basically not applicable to our online setting, it is enough to compare the performance of temporal smoothness.
As a baseline, we use our feature anticipation network when $l_g=12$.
We compare two NMS methods with different windows;
One is using a 5-frame window ranging from $i-2$ to $i+2$-th frames, and the other is using only past information that includes $i-2$ to $i$-th frames.
As shown in Table~\ref{tab:ablation_SP_MI_tables}, the NMS with 5-frame window shows the best performance boost (+1.2\%) by exploiting future information.
However, the performance of the NMS without future information is worse (-0.3\%).
As another baseline, we fixed the weights of PFA module as uniform values (0.5,0.5).
It shows similar performance to that of the NMS without future information (-0.1\%).
To show what is learned on PFA module, we selected an example case in Fig.~\ref{fig:ablation_smoothweight}.
Mostly, PFA outputs high weight for the current frame, but it also outputs high weight for past frames on the predicted action boundaries.
This leads to higher confidence score on action boundaries, and boosts the performance.
Our proposed smooth prediction (SP) shows performance gain (+1.0\%) using only past frames, which is competitive to the NMS with 5-frame window.

\subsubsection{Influence of Multiple Inference}
We verify the influence of multiple inference by running experiments by changing the number of inferences, which is shown in Table~\ref{tab:ablation_SP_MI_tables}.
As the number of inferences increases, the performance also goes up. 
Considering the trade-off between the performance and the computation, we chose the 4-step inference as our default hyperparameter in all the experiments.

\begin{table}[t]
\begin{minipage}[b]{0.59\linewidth}
    \centering
    \begin{tabular}{p{0.7\linewidth}P{0.2\linewidth}}
    \toprule
    Smoothing Method     & mAP \\ \midrule \midrule 
    ATCP             & 48.6   \\
    ATCP + NMS (5-frame window)     & 49.8   \\
    ATCP + NMS (past 3 frames)         & 48.3  \\
    ATCP + SP (Uniform weights)          & 48.5 \\
    ATCP + SP (Ours)         & 49.6  \\ 
    \bottomrule
    \end{tabular}
\end{minipage}
\hfill
\begin{minipage}[b]{0.39\linewidth}
    \centering
    \begin{tabular}{p{0.2\linewidth}P{0.2\linewidth}}
    \toprule
    Steps     & mAP \\ \midrule \midrule
    1       & 49.6   \\
    2       & 51.2   \\
    4       & 51.6  \\
    8       & 51.7   \\
    16      & 51.7    \\ \bottomrule
    \end{tabular}
\end{minipage}
\caption{
    \textit{(left)}
    mAPs on different temporal smoothing methods when $l_g=12$. 
    \textit{(right)}
    mAPs on different steps of multiple inference when $l_g=8$ on ATCP + SP model.
    }
\label{tab:ablation_SP_MI_tables}
\end{table}

\section{Conclusion} \label{Sec:Conclusion}
We have introduced a novel RNN-based network, Future Anticipation and Temporally Smoothing network (FATSnet), for the online action detection task.
FATSnet generates the future information before making a decision, and the anticipation is trained by an unsupervised method with a novel cycle-consistency loss function.
In addition, FATSnet contains the novel temporal smoothing architecture that can be applied in the online setting.
Lastly, FATSnet can minimize the performance loss when an input sequence with excessive length passes through RNN-based models.
We believe our future generation method can give a new intuitions to other online action detection works.
Besides the online action detection task, we believe our approach can serve as a basis for future works on online streaming video understanding tasks, especially on how to handle RNN-based models.

One limitation of our work is that the trained features by cycle-consistency loss are completely black-boxed and it is difficult to provide further interpretation. 
Due to the architectural restriction, the learned features cannot be restored as RGB or other visible formats.
In the future work, we plan to design a visible yet efficient model where learned features can be easily restored to visible formats, and show the inspiring study for generating future frames with the cycle-consistency.

\section*{Acknowledgements}
This work was partially supported by the ICT R\&D program of MSIT/IITP (2017-0-01772, Development of QA systems for Video Story Understanding to pass the Video Turing Test).
Also, this work was supported in part by the Institute of Information and Communications Technology Planning and Evaluation (IITP) Grant funded by the Korean Government (MSIT), Artificial Intelligence Graduate School Program, Yonsei University, under Grant 2020-0-01361.

\bibliography{mybibfile}

\begin{thebibliography}{10}
\expandafter\ifx\csname url\endcsname\relax
  \def\url#1{\texttt{#1}}\fi
\expandafter\ifx\csname urlprefix\endcsname\relax\def\urlprefix{URL }\fi
\expandafter\ifx\csname href\endcsname\relax
  \def\href#1#2{#2} \def\path#1{#1}\fi

\bibitem{tran2015learning}
D.~Tran, L.~Bourdev, R.~Fergus, L.~Torresani, M.~Paluri, Learning
  spatiotemporal features with 3d convolutional networks, in: Proceedings of
  the IEEE International Conference on Computer Vision (ICCV), 2015, pp.
  4489--4497.

\bibitem{li2020spatio}
J.~Li, X.~Liu, M.~Zhang, D.~Wang, Spatio-temporal deformable 3d convnets with
  attention for action recognition, Pattern Recognition 98 (2020) 107037.

\bibitem{hao2019spatiotemporal}
W.~Hao, Z.~Zhang, Spatiotemporal distilled dense-connectivity network for video
  action recognition, Pattern Recognition 92 (2019) 13 -- 24.

\bibitem{chao2018rethinking}
Y.-W. Chao, S.~Vijayanarasimhan, B.~Seybold, D.~A. Ross, J.~Deng,
  R.~Sukthankar, Rethinking the faster r-cnn architecture for temporal action
  localization, in: Proceedings of the IEEE Conference on Computer Vision and
  Pattern Recognition (CVPR), 2018, pp. 1130--1139.

\bibitem{long2019gaussian}
F.~Long, T.~Yao, Z.~Qiu, X.~Tian, J.~Luo, T.~Mei, Gaussian temporal awareness
  networks for action localization, in: Proceedings of the IEEE Conference on
  Computer Vision and Pattern Recognition (CVPR), 2019, pp. 344--353.

\bibitem{gao2020play}
L.~Gao, T.~Li, J.~Song, Z.~Zhao, H.~T. Shen, Play and rewind: Context-aware
  video temporal action proposals, Pattern Recognition 107 (2020) 107477.

\bibitem{paul2018wtalc}
S.~Paul, S.~Roy, A.~K. Roy-Chowdhury, W-talc: Weakly-supervised temporal
  activity localization and classification, in: Proceedings of the European
  Conference on Computer Vision (ECCV), 2018, pp. 563--579.

\bibitem{shou2018autoloc}
Z.~Shou, H.~Gao, L.~Zhang, K.~Miyazawa, S.-F. Chang, Autoloc: Weakly-supervised
  temporal action localization in untrimmed videos, in: Proceedings of the
  European Conference on Computer Vision (ECCV), 2018, pp. 154--171.

\bibitem{de2016online}
R.~De~Geest, E.~Gavves, A.~Ghodrati, Z.~Li, C.~Snoek, T.~Tuytelaars, Online
  action detection, in: Proceedings of the European Conference on Computer
  Vision (ECCV), 2016, pp. 269--284.

\bibitem{hoai2012max}
M.~Hoai, F.~De~la Torre, Max-margin early event detectors, in: Proceedings of
  the IEEE Conference on Computer Vision and Pattern Recognition (CVPR), 2012,
  pp. 2863--2870.

\bibitem{hoai2014max}
M.~Hoai, F.~De~la Torre, Max-margin early event detectors, IJCV 107~(2) (2014)
  191--202.

\bibitem{gao2019startnet}
M.~Gao, M.~Xu, L.~S. Davis, R.~Socher, C.~Xiong, Startnet: Online detection of
  action start in untrimmed videos, in: Proceedings of the IEEE International
  Conference on Computer Vision (ICCV), 2019, pp. 5542--5551.

\bibitem{shou2018online}
Z.~Shou, J.~Pan, J.~Chan, K.~Miyazawa, H.~Mansour, A.~Vetro, X.~Giro-i Nieto,
  S.-F. Chang, Online detection of action start in untrimmed, streaming videos,
  in: Proceedings of the European Conference on Computer Vision (ECCV), 2018,
  pp. 534--551.

\bibitem{THUMOS14}
Y.-G. Jiang, J.~Liu, A.~Roshan~Zamir, G.~Toderici, I.~Laptev, M.~Shah,
  R.~Sukthankar, {THUMOS} challenge: Action recognition with a large number of
  classes, \url{http://crcv.ucf.edu/THUMOS14/} (2014).

\bibitem{shim2018teaching}
M.~Shim, Y.~H. Kim, K.~Kim, S.~J. Kim, Teaching machines to understand baseball
  games: Large-scale baseball video database for multiple video understanding
  tasks, in: Proceedings of the European Conference on Computer Vision (ECCV),
  2018, pp. 404--420.

\bibitem{srivastava2015unsupervised}
N.~Srivastava, E.~Mansimov, R.~Salakhudinov, Unsupervised learning of video
  representations using lstms, in: Proceedings of the International Conference
  on Machine Learning (ICMR), 2015, pp. 843--852.

\bibitem{kalchbrenner2017video}
N.~Kalchbrenner, A.~Oord, K.~Simonyan, I.~Danihelka, O.~Vinyals, A.~Graves,
  K.~Kavukcuoglu, Video pixel networks, in: Proceedings of the International
  Conference on Machine Learning (ICMR), 2017, pp. 1771--1779.

\bibitem{vondrick2017generating}
C.~Vondrick, A.~Torralba, Generating the future with adversarial transformers,
  in: Proceedings of the IEEE Conference on Computer Vision and Pattern
  Recognition (CVPR), 2017, pp. 1020--1028.

\bibitem{kwon2019predicting}
Y.-H. Kwon, M.-G. Park, Predicting future frames using retrospective cycle
  {GAN}, in: Proceedings of the IEEE Conference on Computer Vision and Pattern
  Recognition, 2019, pp. 1811--1820.

\bibitem{gao2018im2flow}
R.~Gao, B.~Xiong, K.~Grauman, Im2flow: Motion hallucination from static images
  for action recognition, in: Proceedings of the IEEE Conference on Computer
  Vision and Pattern Recognition (CVPR), 2018, pp. 5937--5947.

\bibitem{rodriguez2018action}
C.~Rodriguez, B.~Fernando, H.~Li, Action anticipation by predicting future
  dynamic images, in: The European Conference on Computer Vision (ECCV)
  Workshops, 2018.

\bibitem{chaabane2020looking}
M.~Chaabane, A.~Trabelsi, N.~Blanchard, R.~Beveridge, Looking ahead:
  Anticipating pedestrians crossing with future frames prediction, in:
  Preceedings of the IEEE Winter Conference on Applications of Computer Vision
  (WACV), 2020, pp. 2297--2306.

\bibitem{luc2017predicting}
P.~Luc, N.~Neverova, C.~Couprie, J.~Verbeek, Y.~LeCun, Predicting deeper into
  the future of semantic segmentation, in: Proceedings of the IEEE
  International Conference on Computer Vision (ICCV), 2017, pp. 648--657.

\bibitem{vondrick2016anticipating}
C.~Vondrick, H.~Pirsiavash, A.~Torralba, Anticipating visual representations
  from unlabeled video, in: Proceedings of the IEEE Conference on Computer
  Vision and Pattern Recognition (CVPR), 2016, pp. 98--106.

\bibitem{wang2019early}
D.~Wang, Y.~Yuan, Q.~Wang, Early action prediction with generative adversarial
  networks, IEEE Access 7 (2019) 35795--35804.

\bibitem{gammulle2019predicting}
H.~Gammulle, S.~Denman, S.~Sridharan, C.~Fookes, Predicting the future: A
  jointly learnt model for action anticipation, in: Proceedings of the IEEE
  International Conference on Computer Vision (ICCV), 2019, pp. 5562--5571.

\bibitem{de2018modeling}
R.~De~Geest, T.~Tuytelaars, Modeling temporal structure with lstm for online
  action detection, in: Proceedings of the IEEE Winter Conference on
  Applications of Computer Vision (WACV), 2018, pp. 1549--1557.

\bibitem{gao2017red}
J.~Gao, Z.~Yang, R.~Nevatia, {RED}: Reinforced encoder-decoder networks for
  action anticipation, in: Proceedings of the British Machine Vision Conference
  (BMVC), 2017.

\bibitem{xu2019temporal}
M.~Xu, M.~Gao, Y.-T. Chen, L.~S. Davis, D.~J. Crandall, Temporal recurrent
  networks for online action detection, in: Proceedings of the IEEE
  International Conference on Computer Vision (ICCV), 2019, pp. 5532--5541.

\bibitem{xiong2016cuhk}
Y.~Xiong, L.~Wang, Z.~Wang, B.~Zhang, H.~Song, W.~Li, D.~Lin, Y.~Qiao, L.~V.
  Gool, X.~Tang, {CUHK \& ETHZ \& SIAT} submission to activitynet challenge
  2016, arXiv:1608.00797.

\bibitem{caba2015activitynet}
B.~G. Fabian Caba~Heilbron, Victor~Escorcia, J.~C. Niebles, Activitynet: A
  large-scale video benchmark for human activity understanding, in: Proceedings
  of the IEEE Conference on Computer Vision and Pattern Recognition (CVPR),
  2015, pp. 961--970.

\bibitem{zhu2017unpaired}
J.-Y. Zhu, T.~Park, P.~Isola, A.~A. Efros, Unpaired image-to-image translation
  using cycle-consistent adversarial networks, in: Proceedings of the IEEE
  International Conference on Computer Vision (ICCV), 2017, pp. 2223--2232.

\bibitem{wang2016temporal}
L.~Wang, Y.~Xiong, Z.~Wang, Y.~Qiao, D.~Lin, X.~Tang, L.~Van~Gool, Temporal
  segment networks: Towards good practices for deep action recognition, in:
  Proceedings of the European Conference on Computer Vision (ECCV), 2016, pp.
  20--36.

\bibitem{carreira2017quo}
J.~Carreira, A.~Zisserman, Quo vadis, action recognition? a new model and the
  kinetics dataset, in: Proceedings of the IEEE Conference on Computer Vision
  and Pattern Recognition (CVPR), 2017, pp. 6299--6308.

\bibitem{he2016deep}
K.~He, X.~Zhang, S.~Ren, J.~Sun, Deep residual learning for image recognition,
  in: Proceedings of the IEEE Conference on Computer Vision and Pattern
  Recognition (CVPR), 2016, pp. 770--778.

\bibitem{kingma2014adam}
D.~P. Kingma, J.~Ba, Adam: A method for stochastic optimization, arXiv preprint
  arXiv:1412.6980.

\bibitem{simonyan2014very}
K.~Simonyan, A.~Zisserman, Very deep convolutional networks for large-scale
  image recognition, arXiv:1409.1556.

\bibitem{simonyan2014two}
K.~Simonyan, A.~Zisserman, Two-stream convolutional networks for action
  recognition in videos, in: Proceedings of the Advances in neural information
  processing systems (NeurIPS), 2014, pp. 568--576.

\bibitem{shou2017cdc}
Z.~Shou, J.~Chan, A.~Zareian, K.~Miyazawa, S.-F. Chang, {CDC}:
  Convolutional-de-convolutional networks for precise temporal action
  localization in untrimmed videos, in: Proceedings of the IEEE conference on
  computer vision and pattern recognition (CVPR), 2017, pp. 5734--5743.

\bibitem{yeung2018every}
S.~Yeung, O.~Russakovsky, N.~Jin, M.~Andriluka, G.~Mori, L.~Fei-Fei, Every
  moment counts: Dense detailed labeling of actions in complex videos,
  International Journal of Computer Vision (IJCV) 126~(2-4) (2018) 375--389.

\end{thebibliography}

\end{document}